\documentclass[10pt,twocolumn,letterpaper]{article}

\usepackage[pagenumbers]{cvpr} %

\usepackage[dvipsnames]{xcolor}

\makeatletter
\def\adl@drawiv#1#2#3{%
        \hskip.5\tabcolsep
        \xleaders#3{#2.5\@tempdimb #1{1}#2.5\@tempdimb}%
                #2\z@ plus1fil minus1fil\relax
        \hskip.5\tabcolsep}
\newcommand{\cdashlinelr}[1]{%
  \noalign{\vskip\aboverulesep
           \global\let\@dashdrawstore\adl@draw
           \global\let\adl@draw\adl@drawiv}
  \cdashline{#1}
  \noalign{\global\let\adl@draw\@dashdrawstore
           \vskip\belowrulesep}}
\makeatother

\usepackage{booktabs}       %
\usepackage{amsfonts}       %
\usepackage{nicefrac}       %
\usepackage{microtype}      %
\usepackage{bbding}
\usepackage{times}
\usepackage{epsfig}
\usepackage{graphicx}
\usepackage{tabularx}
\usepackage{amsmath}
\usepackage{amssymb}
\usepackage{arydshln}
\usepackage{multirow}
\usepackage{cuted}
\usepackage{capt-of} %

\newcommand\lft{\mathopen{}\left}
\newcommand\rgt{\aftergroup\mathclose\aftergroup{\aftergroup}\right}

\definecolor{Gray}{gray}{0.92}
\usepackage{colortbl}

\newcommand{\supparxiv}[2]{#2}
\newcommand{\mypar}[1]{\vspace{-3.5mm}\paragraph{#1}}
\def\upvspacefig{\vspace{-0mm}}
\def\projecturl{https://cfeng16.github.io/UniTouch/}

\definecolor{cvprblue}{rgb}{0.21,0.49,0.74}
\definecolor{urlblue}{rgb}{0.24,0.49,1.0}
\usepackage[pagebackref,breaklinks,colorlinks,citecolor=cvprblue]{hyperref}
\usepackage[export]{adjustbox} %
\usepackage{overpic}

\def\paperID{533} %
\def\confName{CVPR}
\def\confYear{2024}

\title{Binding Touch to Everything: \\Learning Unified Multimodal Tactile Representations}

\author{Fengyu Yang\textsuperscript{1*} \quad
Chao Feng\textsuperscript{2*} \quad
Ziyang Chen\textsuperscript{2*} \quad
Hyoungseob Park\textsuperscript{1} \quad
Daniel Wang\textsuperscript{1} \quad
Yiming Dou\textsuperscript{2} \vspace{1.1mm} \\
Ziyao Zeng\textsuperscript{1} \quad
Xien Chen\textsuperscript{1} \quad
Rit Gangopadhyay\textsuperscript{1} \quad
Andrew Owens\textsuperscript{2} \quad
Alex Wong\textsuperscript{1}
\vspace{3mm} \\
\textsuperscript{1}Yale University \quad \textsuperscript{2}University of Michigan \\
}

\begin{document}
\maketitle
{\let\thefootnote\relax\footnotetext{{{*} Indicates equal contribution.}}}

\begin{strip}
\centering
    \centering
    \raggedright
    \vspace{-11mm} %
    \includegraphics[width=\textwidth]{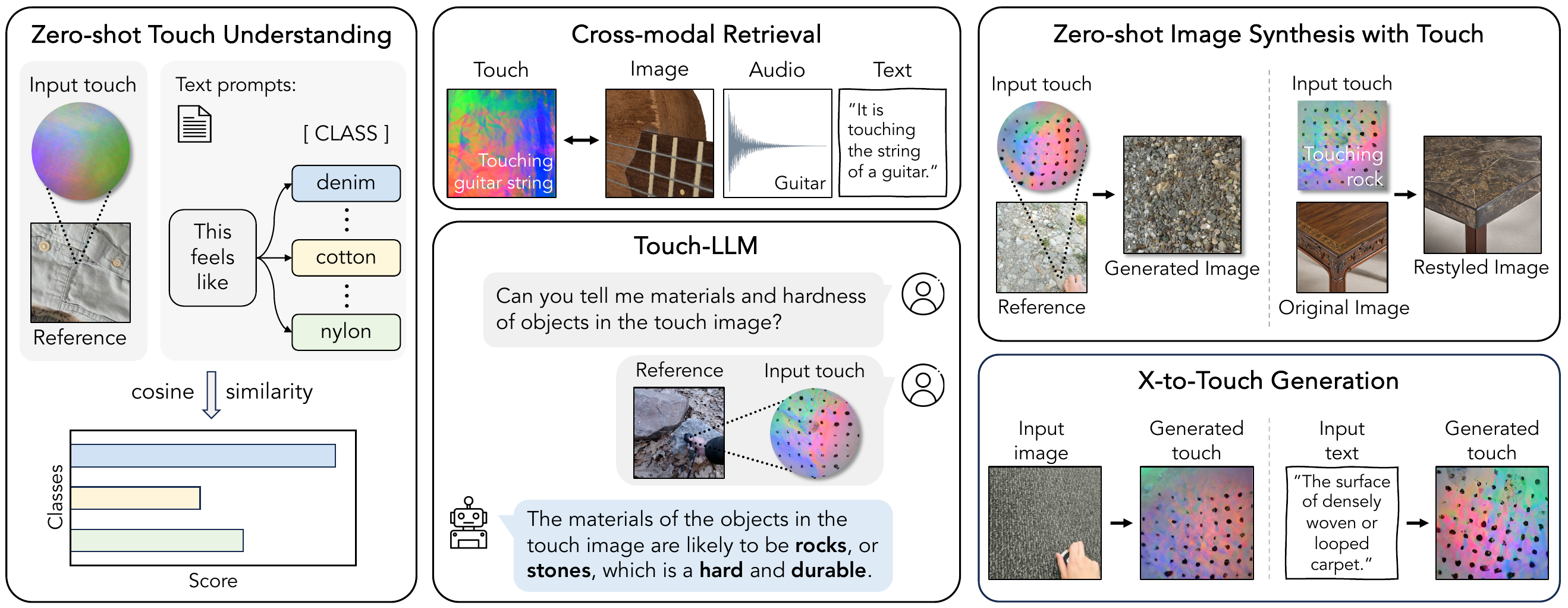}

      \captionof{figure}{\textbf{Putting touch ``in touch'' with other modalities.} We show that a variety of tactile sensing tasks, ranging from touch image understanding to image synthesis with touch, can be solved zero-shot by aligning touch to pretrained multimodal models, extending previous approaches on work on other modalities~\cite{Girdhar2023ImageBindOE}. Our learned model can be applied to various vision-based tactile sensors and simulators (\eg, GelSight, DIGIT, Taxim, and Tacto). For visualization purposes, we show the corresponding visual signal (labeled ``reference'') for each touch signal, even though it is not used by the model. }%

  \vspace{3mm}
\label{fig:teaser}
\end{strip}
\begin{abstract}
\vspace{-2mm}
The ability to associate touch with other modalities has huge implications for humans and computational systems. However, multimodal learning with touch remains challenging due to the expensive data collection process and non-standardized sensor outputs. We introduce UniTouch, a unified tactile model for vision-based touch sensors connected to multiple modalities, including vision, language, and sound. We achieve this by aligning our UniTouch embeddings to pretrained image embeddings already associated with a variety of other modalities. We further propose learnable sensor-specific tokens, allowing the model to learn from a set of heterogeneous tactile sensors, all at the same time. UniTouch is capable of conducting various touch sensing tasks in the zero-shot setting, from robot grasping prediction to touch image question answering. To the best of our knowledge, UniTouch is the first to demonstrate such capabilities. Project Page: \small{\url{\projecturl}}.

\end{abstract}    
\section{Introduction}
\label{sec:intro}

Amongst our five main senses, touch sensing is perhaps the most crucial to human survival, due to its role in perceiving physical contact  --- rivaling even vision in its  overall importance~\cite{hutmacher2019there,manske1999sense,linden2016touch}. 
Our ability to form cross-modal associations between touch and our other senses~\cite{smith2005development} thus underlies a great deal of our physical capabilities. For example, we predict from vision how a surface will feel before we touch it, and we predict from touch how an object will sound before we strike it. These cross-modal associations are also a key component of computational systems, such as for robotic manipulation~\cite{feel_success, Li2023ViHOPE, Qi2023GeneralIO, Xu2021TowardsLT, touch-dexterity, Lloyd2020GoalDrivenRP, lepert2023inhand, Pan2022InHandMO, yu2022fully, Cao2021MultimodalPF, Pecyna2022VisualTactileMF}, material and geometry estimation~\cite{yang2022touch, yuan2017shape, gupta2021tactile, Cao2021TouchRollerAR}, assistive technology~\cite{higuera2023learning}, and texture recognition~\cite{Yuan2017ConnectingLA, Luo2018ViTacFS, Jiang2021RoboticPO}.

\begin{figure}[!t]
\centering
\upvspacefig
\scriptsize

\setlength{\tabcolsep}{5pt}
 \renewcommand{\arraystretch}{1.4}
\begin{tabular}{ccc}
GelSight from \cite{yang2022touch}     &    
DIGIT from \cite{suresh2022midastouch}   & 
Taxim from \cite{gao2022ObjectFolderV2} \\
\frame{\includegraphics[width=0.3\linewidth]{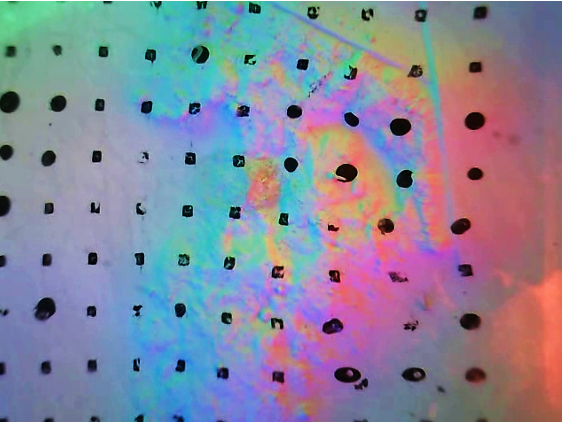}} &
\frame{\includegraphics[width=0.3\linewidth]{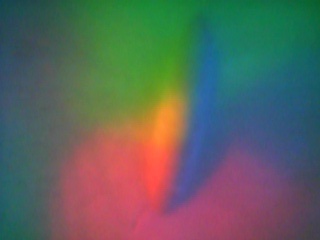}} &
\frame{\includegraphics[width=0.3\linewidth]{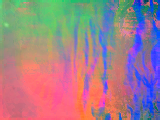}}  \\

GelSlim from \cite{Gao_2023_CVPR}     &    
TACTO from \cite{gao2021ObjectFolder}   & 
DIGIT from \cite{kerr2022ssvtp} \\
\frame{\includegraphics[width=0.3\linewidth]{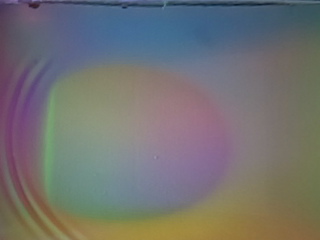}} &
\frame{\includegraphics[width=0.3\linewidth]{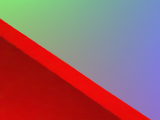}} &
\frame{\includegraphics[width=0.3\linewidth]{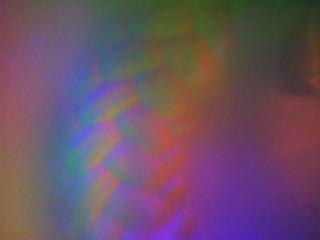}} \\

\end{tabular}
    
\caption{{\bf Tactile images of different sensors and datasets.} In contrast to many other modalities, signals from different touch sensing hardware exhibit large amounts of variation. 
}
\label{fig:sensor}
\end{figure}

Despite their importance, cross-modal associations between touch and other modalities have received considerably less attention from the multimodal research community than those of other modalities, such as vision, language, and sound. 
Touch is expensive to acquire~\cite{yang2022touch, gao2021ObjectFolder, gao2022ObjectFolderV2}
as it requires actively probing objects with touch sensors, limiting the scale of data collected for training tactile ``foundation'' models. Moreover, touch sensors are not fully standardized, and thus there are large differences between outputs of different sensors~\cite{Gao2021OnEA, Zandonati2023InvestigatingVF}. Even amongst the commonly used vision-based sensors, the difference in mechanical design and elastomeric material will lead to divergent artifacts, limiting generalization (\cref{fig:sensor}). As a result, existing tactile representations are typically constrained to a single sensor.

An emerging line of work has addressed the challenges of learning from other low-resource modalities, like sound, point clouds, and depth, by aligning examples with pretrained vision-language embeddings~\cite{lee2022sound,Girdhar2023ImageBindOE, Xue2022ULIPLA}. %
In this paper, we show that this approach can be adapted to tactile sensing. We align tactile signals to visual signals, thereby linking touch to a variety of other modalities, such as language and sound. 
Then we can use the representations within off-the-shelf models trained on other modalities (\eg CLIP~\cite{Radford2021LearningTV}), to solve a variety of tactile sensing tasks. To deal with the large variations in different touch sensors, we train a single model with multiple tactile signals at once, and introduce learnable tokens to model sensor-specific properties, such as the calibration and intensity profiles in the touch signal.

Our trained model, which we call \textbf{UniTouch}, is a general-purpose interface for multiple vision-based tactile sensors. Our model unifies many previously studied tactile sensing tasks ``zero shot" and greatly expands the range of tasks that touch sensing can be applied, as shown in \cref{fig:teaser}:
(i) We apply it to zero-shot touch understanding tasks like material recognition and robotic grasp stability prediction. (ii) We obtain strong performance in cross-modal retrieval with touch by aligning touch with other modalities in a shared latent space. (iii) The learned representation can also support image synthesis tasks, including touch-to-image generation~\cite{Li2019ConnectingTA, yang2023generating} and tactile-driven image stylization~\cite{yang2022touch, yang2023generating}, by using it within off-the-shelf text-to-image diffusion models. 
(iv) We combine touch with large language models (LLM),
allowing us to perform tasks such as tactile question answering in a variety of tactile domains, including contact localization, grasping stability prediction, and \etc. (v) Finally, we perform ``X-to-touch'' generation, producing touch images from vision, text, and audio. Our experiments suggest our zero-shot model achieves competitive (or even better) performance than previously proposed approaches on multiple tasks.

\section{Related Work}
\label{sec:related_work}
\vspace{3mm}
\mypar{Tactile sensing.}
Early tactile sensors were chiefly engineered to register fundamental, low-dimensional sensory outputs such as force, pressure, vibration, and temperature~\cite{Lederman1987HandMA, Lederman2009TUTORIALRH, Cutkosky2008ForceAT, Kappasov2015TactileSI}. 
Lately, there has been a growing focus on vision-based tactile sensors. GelSight ~\cite{Yuan2017GelSightHR,johnson2009retrographic} as one of the representative sensors, features an elastomeric gel with an embedded camera and illumination system. The gel deforms upon contact with an object and creates a high-resolution height map using photometric stereo~\cite{Johnson2011MicrogeometryCU}, which provides detailed information about the shape and physical properties of touch~\cite{Taylor2021GelSlim3H, Lepora2022DigiTacAD}. 
One variant, DIGIT~\cite{Lambeta2020DIGITAN}, has a specially designed silicone-based elastomer gel with a harder surface and a different illumination system. Another variant GelSlim~\cite{Taylor2021GelSlim3H} contains a stretchy, loose-weave fabric gel surface. Recent work also turns into the simulation of tactile sensors~\cite{Si2021TaximAE, Wang2020TACTOAF, Agarwal2020SimulationOV, Gomes2023BeyondFG, Church2021TactileSP, Jianu2021ReducingTS}. Taxim~\cite{Si2021TaximAE} simulates the optical response of a GelSight sensor and TACTO~\cite{Wang2020TACTOAF} calculates the local contact geometry and the corresponding rendering. 
We focus on these vision-based sensors as they are widely available in visuo-tactile datasets~\cite{gao2021ObjectFolder, gao2022ObjectFolderV2, Gao_2023_CVPR, Yuan2017ConnectingLA, suresh2022midastouch, Yuan2017GelSightHR, yang2022touch, Wang2019RealTimeSB, Xu2023VisualTactileSF}, are commonly used in various applications~\cite{Li2023ViHOPE, Gao_2023_ICCV, Suresh22icra, Chaudhury2022UsingCV, Lambeta2021PyTouchAM, zhong2023touching, Huang2022UnderstandingDT, Li2022SeeHA, Heravi2019LearningAA, Yu2023MimicTouchLH, Cao2020SpatiotemporalAM, Jiang2021VisionGuidedAT,Cao2023Vis2HapVH}, and all adopt image as the output format.
While these vision-based tactile sensors and simulators share similar imaging patterns, the difference in design and calibration results in a significant domain gap, as shown in \cref{fig:sensor}.
Hence, researchers typically study each sensor separately. In our work, we introduce a novel approach to understanding multiple sensors through our unified touch encoder.

\mypar{Representation learning with touch.}
The initial efforts learn tactile representations for specific tasks~\cite{Gao2020SupervisedAJ, Taunyazov2020FastTC, Lee2019MakingSO, Yuan2017ConnectingLA, Lin2019LearningTI}. 
Lee \etal~\cite{Lee2019MakingSO} undertook a collaborative training of Convolutional Neural Networks (CNN) for an RGB camera and a force sensor to facilitate contact-rich manipulation tasks. Similarly, Yuan \etal~\cite{Yuan2017ConnectingLA} employed a comparable methodology to establish a shared latent space between visual and tactile modalities using the Gelsight touch sensor, aimed at precise fabric classification. 
Recently, researchers have learned general representations of touch through self-supervision.
Yang \etal~\cite{yang2022touch} learned tactile representations for Gelsight sensors with visuo-tactile contrastive multiview coding~\cite{tian2020contrastive} and Kerr \etal~\cite{kerr2022ssvtp} proposed a contrastive pretraining method for the DIGIT sensor. 
Other works adopted BYOL framework~\cite{guzey2023dexterity} or contrastive predictive coding~\cite{Zambelli2021LearningRT} to learn representations for non vision-based tactile sensors like BioTac. 
Some work~\cite{jiang2023learn} applies masked autoencoders to learn tactile representations directly from tactile inputs. 
Unlike methods concentrated solely on visuo-tactile learning for a single sensor, our approach aims to learn touch representations that can be applied across various sensors and interconnected with multiple modalities.

\mypar{Multimodal representation learning.} 
The success of vision-language pretraining ~\cite{desai2021virtex, radford2021learning, xu2021videoclip, luo2022clip4clip, qiu2021vt} has demonstrated the ability to bridge the gap between visual content, such as images or videos, and textual descriptions~\cite{ji2023mrtnet, ji2023online, li2023dynamic}. 
Furthermore, some researchers have extended the application of CLIP into the 3D domain~\cite{zhang2021pointclip, Zhu2022PointCLIPV2, DepthCLIP, guo2023point}. %
Some works learn shared audio-visual representation~\cite{owens2018ambient, asano2020labelling, morgado2021audio, hu2022mix, iQuery, xu2022ava, du2023conditional, sungbin2023sound, feng2023self} by leveraging natural correspondence with videos. 
Some works also study shared audio-language representation~\cite{guzhov2022audioclip, wu2023large, elizalde2023clap}. 
Bender \etal~\cite{bender2023learning} crafted an embedding space for the flavors of wines by leveraging both image and text annotations.
Chen \etal~\cite{chen2023sound} learned shared spatial information from binaural sound and vision. 
Some works learned the association between vision and metadata~\cite{zheng2023exif, wu2021mgh,chen2023movies2scenes}. %
Imagebind~\cite{Girdhar2023ImageBindOE} proposed to learn a joint embedding for six diverse modalities solely through image alignment and emerge zero-shot cross-modal capabilities. 
In our work, we extend this concept to the sense of touch and bind it to other modalities including text and audio by aligning tactile data with images, encouraging a more comprehensive understanding of cross-modal touch interactions without paired data.

\section{Method}

\newcommand{\bv}{{\mathbf v}}
\newcommand{\bt}{{\mathbf t}}
\newcommand{\bs}{{\mathbf s}}
\newcommand{\ft}{\mathcal{F}_T}
\newcommand{\fv}{\mathcal{F}_V}

\begin{figure}[!t]
    \centering
    \upvspacefig
    \includegraphics[width=\linewidth]{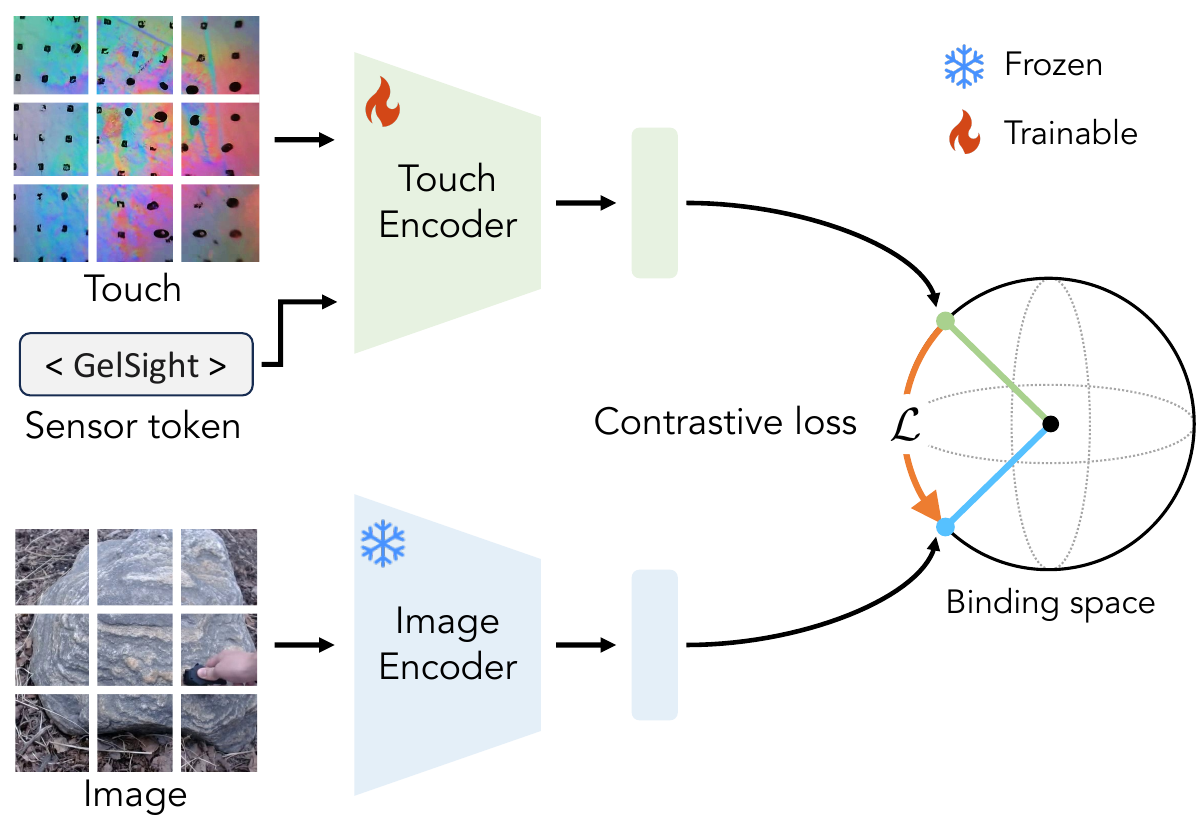}
    
    \caption{{\bf Method overview.}  We align our touch embedding with a pre-trained image embedding derived from large-scale vision language data, using sensor-specific tokens for multi-sensor training.  } 
    \label{fig:method_overview}
\end{figure}

We aim to learn a unified tactile representation for different touch sensors that captures relationships between touch and different modalities, \eg. vision, text, and audio. 
First, we present our contrastive visuo-tactile pretraining, inspired by \cite{Girdhar2023ImageBindOE}, that can emerge interconnections of touch and other modalities. We then introduce our touch encoder design and data sampling strategy that can be used for different tactile sensors at once. Finally, we show how our learned representation can be applied to various downstream tasks.

\subsection{Binding touch with images}
\label{sec:touchbind}
We learn a multimodal tactile representation from touch and vision solely, without the need for paired text and audio data for touch. 
We achieve that by aligning our touch embedding to a pretrained image embedding using contrastive learning as shown in \cref{fig:method_overview}, where the image embedding is already aligned with modalities like language and audio training from large-scale image-paired datasets~\cite{Girdhar2023ImageBindOE}.

We denote $\Omega_v$ as the visual image domain and $\Omega_t$ as the tactile image domain. Thus, given $B$ visual and touch pairs in a batch, $\{(\bv_i, \bt_i)\}_{i=1}^B$, where $\bv_i: \Omega_v \subset \mathbb{R}^2 \rightarrow \mathbb{R}^3$ and  $\bt_i: \Omega_t \subset \mathbb{R}^2 \rightarrow \mathbb{R}^3$, we align a tactile embedding $\ft(\bt_i) \in  \mathbb{R}^C$ with the pretrained visual embedding $\fv(\bv_i) \in \mathbb{R}^C$ from \cite{Girdhar2023ImageBindOE} by maximizing the cosine similarity between corresponding visuo-tactile pairs. We optimize this objective using InfoNCE loss~\cite{oord2018representation} to match touches to correct images: 
\vspace{-2mm}
\begin{equation}\label{eq1}
  \mathcal{L}_{T \rightarrow V} = -\frac{1}{B}\sum_{i = 1}^{B}{\log}\frac{\exp(\ft(\bt_i) \cdot \fv(\bv_i)/\tau)}
  {\sum_{j = 1}^{B} {\exp}\lft(\ft(\bt_i\rgt) \cdot \fv\lft(\bv_j)/\tau\rgt)} \text{,}
\end{equation}
where $\tau$ is a temperature hyperparameter~\cite{wu2018unsupervised} and $C$ is feature dimension. Analogously, we can also match from image $\bv_i$ to touch $\bt_i$ using the loss $\mathcal{L}_{V \rightarrow T}$. Thus, we minimize the overall loss: 
\begin{equation}\label{eq2}
\mathcal{L} = \mathcal{L}_{T \rightarrow V} + \mathcal{L}_{V \rightarrow T} \text{.}
\end{equation}

Naturally, minimizing the contrastive objective~\cite{feng2023self, zheng2023exif, yang2022sparse, tian2020contrastive} will ``pull'' a visuo-tactile pair close together and ``push'' it away from other pairs, achieving the alignment between touch and visual embedding. 
As the visual embedding comes from a learned joint space that has already aligned with different modalities, touch that is bound with images will bridge a connection to other modalities, yielding a multi-modal unified tactile representation.

\subsection{Learning from multiple sensors at once}
\label{sec:touchnet}

We want to learn a generalizable tactile representation that will be suitable for different tactile sensors. Therefore, we designed our touch encoder $\ft$ to bridge the domain gap among various vision-based tactile sensors caused by the difference in sensor designs. 

Specifically, we introduce a set of learnable sensor-specific tokens $\{\bs_k \}_{k=1}^K$, where $\bs_k \in R^{L \times D}$, to capture specific details for each senor, \eg, calibration and background color in touch images, so that the remaining model capacity can be used to learn common knowledge across different type of touch sensors, such as texture and geometry. 
Here, $K$ represents the number of sensors we train on, $L$ is the number of sensor-specific tokens for each sensor, and $D$ is the token dimension.
For the given touch image $\bt_i$, and its corresponding tactile sensor tokens $\bs_{\bt_i}$, we append these sensor-specific tokens as prefixes to touch image patch tokens and then encode them with our touch encoder resulting in the final embedding $\ft(\bt_i, \bs_{\bt_i})$ (\cref{fig:method_overview}). For our contrastive vision-touch pretraining, we optimize: 
\vspace{-1mm}
\begin{equation}\label{eq3}
  \mathcal{L}_{T \rightarrow V} = - \frac{1}{B}\sum_{i = 1}^{B}{\log}\frac{\exp(\ft(\bt_i, \bs_{\bt_i}) \cdot \fv(\bv_i)/\tau)}
  {\sum_{j = 1}^{B} {\exp}\lft(\ft(\bt_i, \bs_{\bt_i}\rgt) \cdot \fv\lft(\bv_j)/\tau\rgt)} \text{,}
\end{equation}
as well as $\mathcal{L}_{V \rightarrow T}$ from the other direction.

\mypar{In-batch data sampling.}
We found that batch sampling strategy~\cite{Cui2021ContrastiveVP} plays an important role when we train with data, acquired by multiple touch sensors, using contrastive learning. 
The model will under-perform if we randomly sample from each data source~\cite{Yang2022UnifiedCL} which results in a surplus of easy negatives due to the domain gap between different sensors. 
Therefore, we design a batch sampling strategy to guarantee that $\sigma$ percent of training examples in a batch are sampled from the same datasets. Given that our dataset $\mathcal{D}$ is the union over $N$ datasets collected with diverse tactile sensors $\mathcal{D} = \bigcup_{n \in \{1, 2, ..., N\}} \mathcal{D}_n$, the probability of selecting a given dataset $D_n$ to sample from is defined as: 
\vspace{-1mm}
\begin{equation}\label{eq4}
  p_n = \frac{\|\mathcal{D}_n\|}{\sum_{m = 1}^{N}\|\mathcal{D}_m\|} \text{,}
  \vspace{-1mm}
\end{equation}
where $\|\cdot\|$ denotes cardinality. $\mathcal{D}_\sigma$ denotes the selected dataset from which we perform uniform random sampling to yield $\sigma \cdot B$ examples;
the rest $(1 -\sigma) \cdot B $ examples are uniformly sampled from other datasets, i.e., $\mathcal{D}  \setminus \mathcal{D}_\sigma$, where $\sigma$ is a hyperparameter range from 0 to 1 representing the portion of the batch.
This batch sampling strategy significantly benefits our training as it allows the model to mostly focus on intra-sensor hard negatives but still be exposed to different sensors to enhance inter-sensor discrimination.

\mypar{Inference.} 
To generalize our learned representation to unseen types of sensors during the inference, we retrieve the nearest neighbor sensor-specific tokens from the learned sensor set $\{\bs_k \}_{k=1}^N$. Specifically, we first compute a prototype for each sensor, a 1D vector that averages all the raw pixels belonging to the tactile images collected by this sensor, and store these prototypes after training. Then, during the inference stage, we compute the L1 distance of between an input tactile image and all the sensor prototypes and retrieve the sensor with minimum distance.

\subsection{Applications}
By aligning our touch embedding to the joint latent space, we establish a link between touch and other modalities. These alignments allow us to perform various zero-shot and cross-modal applications without any further training. 

\mypar{Zero-shot touch understanding.}
Emergent alignment of touch and text enables zero-shot touch understanding, \eg, material classification and grasp stability prediction. Following CLIP~\cite{radford2021learning}, we encode the touch images and text prompts with templates and class names. We compute their similarity score and rank them to achieve the zero-shot classification.

\mypar{Touch-LLM.} 
Using an existing vision-language model~\cite{zhang2023llama,gao2023llama} with the image embedding~\cite{Girdhar2023ImageBindOE} that we align our touch embedding with, we can create our touch-language model by switching to our touch encoder. Given the touch image and language inputs, we can obtain a more comprehensive understanding via question-answering.   

\mypar{Image synthesis with touch.} 
Binding touch with text also opens up more potential abilities for image synthesis with touch. We leverage the pretrained text-to-image diffusion model~\cite{rombach2022high} and use our touch features to condition the denoising process, achieving zero-shot touch-to-image generation~\cite{Li2019ConnectingTA, yang2023generating} and tactile-driven image stylization.

\mypar{X-to-touch generation.} 
We also connect other modalities to touch using the diffusion model so that we can achieve x-to-touch generation, where we imagine the touch by seeing, describing, or listening.
We train an image-to-touch diffusion model~\cite{yang2023generating} using the pretrained joint image embedding and then we can generate touch from text and audio as well. 

\begin{table}[!t]

    \centering
    \renewcommand{\arraystretch}{1.1}
    \resizebox{\linewidth}{!}{
    \setlength{\tabcolsep}{4pt}
    \begin{tabular}{lllccc}
    \toprule
        & \textbf{Dataset}  & \bf Sensor & \textbf{\# data} & \textbf{\begin{tabular}[c]{@{}c@{}}Material\\ cls.\end{tabular}} & \textbf{\begin{tabular}[c]{@{}c@{}}Robot\\ grasp\end{tabular}}\\ \midrule
        \parbox[t]{3mm}{\multirow{4}{*}{\rotatebox[origin=c]{90}{Train \& Eval}}}
         & Touch and Go~\cite{yang2022touch} & GelSight & 120k  & \checkmark & \\
        & The Feeling of Success~\cite{feel_success} & GelSight & 9.3k &   & \checkmark \\
        & YCB-Slide~\cite{suresh2022midastouch} & DIGIT & 183k & \checkmark  &  \\
        & Object Folder 2.0~\cite{gao2022ObjectFolderV2} & Taxim & 180k &  \checkmark & \checkmark \\ 
        \midrule
        \parbox[t]{3mm}{\multirow{3}{*}{\rotatebox[origin=c]{90}{Eval.}}}  & Object Folder Real~\cite{Gao_2023_CVPR} & GelSlim & 20k  &  \checkmark &  \\ 
        & Object Folder 1.0~\cite{gao2021ObjectFolder} & TACTO & 20k & \checkmark  & \checkmark \\ 
         & SSVTP~\cite{kerr2022ssvtp} & DIGIT & 4.6k &  \checkmark &  \\ 
    \bottomrule
    \end{tabular}
    }
     \caption{\textbf{Datasets for training and evaluation.}
    }
    \label{tab:dataset}
\end{table}

\begin{table*}[!t]
\footnotesize
    \upvspacefig

\centering
\resizebox{\linewidth}{!}{
\begin{tabular}{lllccccccc}
\toprule
& \multirow{2}{*}{\textbf{Method}} & \multirow{2}{*}{\shortstack[c]{\textbf{Pretrain} \\ \textbf{Data}}} & \multicolumn{3}{c}{\textbf{In domain Datasets}} & & \multicolumn{3}{c}{\textbf{Out-of-domain Datasets}}\\ 
\cmidrule{4-6} \cmidrule{8-10} 
&  &  & \emph{Touch and Go} & \emph{ObjectFolder 2.0} & \emph{YCB-Slide} & & \emph{ObjectFolder 1.0} & \emph{ObjectFolder Real} & \emph{SSVTP} \\ 
\midrule
 & Chance & -- & 5.0 & 14.2 & 10.0 & & 14.2 & 14.2 & 16.6 \\  
 \cdashlinelr{1-10} 
\multirow{6}{*}{Linear Probing} & Supervised & ImageNet & 47.1 & 70.3& 72.3 & & 37.5 & 54.8 & 73.4 \\
 & VT CMC~\cite{yang2022touch} & Single & 56.5 & 74.3 & 75.2 & & -- & -- & -- \\
 & SSVTP~\cite{kerr2022ssvtp} & Single & 47.6 & 69.8 & 74.8 & & -- & -- & -- \\ 
  & VT CMC~\cite{yang2022touch} & All & 49.2 & 70.3 & 69.5 & & 33.8 & 48.1 & 68.5 \\
 & SSVTP~\cite{kerr2022ssvtp} & All & 43.8 & 68.9 & 67.4 & & 35.1 & 49.7 & 66.8 \\ 
 & Ours& All & \textbf{61.3} & \textbf{85.4} & \textbf{78.1} & & \textbf{41.3} & \textbf{61.2} & \textbf{77.4} \\
 \midrule
 \multirow{1}{*}{Zero-Shot} 
 & Ours& All & 52.7 & 43.5 & 66.4 & & 32.7 & 33.2 & 60.9 \\
\bottomrule
\end{tabular}
}
\caption{\textbf{Tactile material classification.} We compare our touch features with other methods and ImageNet pretraining. We also report our zero-shot classification performance. The metric is accuracy~(\%).}
\label{tab:cls}
\end{table*}

\begin{table}[!t]
\footnotesize

\centering
\resizebox{\linewidth}{!}{
\setlength{\tabcolsep}{0.9mm}
{
\begin{tabular}{lllccc}
\toprule
& \multirow{2}{*}{\textbf{Method}} & \multirow{2}{*}{\shortstack[c]{\textbf{Pretrain} \\ \textbf{Data}}} & \multicolumn{2}{c}{\textbf{In domain}}  & \multicolumn{1}{c}{\textbf{Out-of-domain}}\\ \cmidrule{4-6}
&  &   & \emph{Feeling} & \emph{OF 2.0} & \emph{OF 1.0}\\ \midrule
 & Chance & - & 52.3 & 52.0 & 50.7\\  \cdashlinelr{1-6} 
\multirow{6}{*}{\shortstack[c]{Linear \\ Probing}} & Supervised & ImageNet  & 75.9 & 70.1 & 68.9\\
 & VT CMC~\cite{yang2022touch} & Single& 80.1 & 74.8 & -\\
 & SSVTP~\cite{kerr2022ssvtp} & Single& 80.3 & 74.0 & -\\ 
 & VT CMC~\cite{yang2022touch} & All & 66.1 & 65.8 & 67.2\\
 & SSVTP~\cite{kerr2022ssvtp} & All & 65.8 & 64.2 & 65.3\\ 
 & Ours & All & \textbf{82.3} & \textbf{78.1} & \textbf{75.8}\\
 \midrule
 \multirow{1}{*}{Zero-Shot} 
 & Ours & All & 65.5 & 64.3 & 64.7\\
\bottomrule
\end{tabular}}
}
\caption{ \textbf{Robotics grasping stability prediction.} We compare our touch features with other methods and ImageNet pretraining on grasping stability prediction task. We report our zero-shot results. The metric is accuracy~(\%). }
\label{tab:grasp}
\end{table}

\section{Experiments}
We evaluate our model on extensive tasks spanning various application domains, including zero-shot touch understanding, cross-modal retrieval, zero-shot image synthesis with touch, Touch-LLM, and X-to-touch generation.

\mypar{Implementations.}
We base our model on ImageBind~\cite{Girdhar2023ImageBindOE}. We use the AdamW optimizer~\cite{kingma2015adam,loshchilov2017decoupled} with the base learning rate of $1 \times 10^{-5}$ and cosine decay learning rate scheduler. We train our model with a batch size of 48 on each of the 4 NVIDIA A40 GPUs for 150 epochs. We set the temperature parameter $\tau=0.07$.  We adopt Vision Transformer (ViT)~\cite{dosovitskiy2021an} as the backbone for our touch encoder, which contains 24 multi-head attention blocks with 16 heads on each. The feature dimension $C$ is 1024. 
We use $L = 5$ learnable tokens for each sensor type in our pretraining datasets with $K=3$ different sensors. For the in-batch sampling, we set $\sigma = 0.75$, meaning that 75\% of the data comes from the same dataset, with the remainder sourced from others.

\mypar{Datasets.}
We train and evaluate our model on four visuo-tactile datasets collected by three different vision-based tactile sensors~(\cref{tab:dataset}).
These include the real-world dataset Touch and Go~\cite{yang2022touch}, the robotic dataset Feeling of Success~\cite{feel_success}, the YCB-Slide~\cite{suresh2022midastouch} dataset featuring DIGIT sensor interactions, and the multimodal dataset ObjectFolder 2.0~\cite{gao2022ObjectFolderV2} which contains simulated visual, tactile, and audio data of daily objects using Taxim tactile simulators.
We train our model solely on the naturally paired image and touch data via self-supervision. 
To test the generalization ability of our model, we also evaluate it with three out-of-domain datasets with two unseen sensors, including ObjectFolder Real~\cite{Gao_2023_CVPR}, ObjectFolder 1.0~\cite{gao2021ObjectFolder} and SSVTP~\cite{kerr2022ssvtp}. We specifically select objects 101-1000 from ObjectFolder 2.0 to avoid overlap with ObjectFolder 1.0. Also, ObejctFolder Real contains objects distinct from those in ObjectFolder 1.0 and 2.0.
Please see \supparxiv{the supplement}{\cref{sec:supp_dataset}} for more details.

\begin{figure*}[t]
    \upvspacefig

    \centering
    \includegraphics[width=\linewidth]{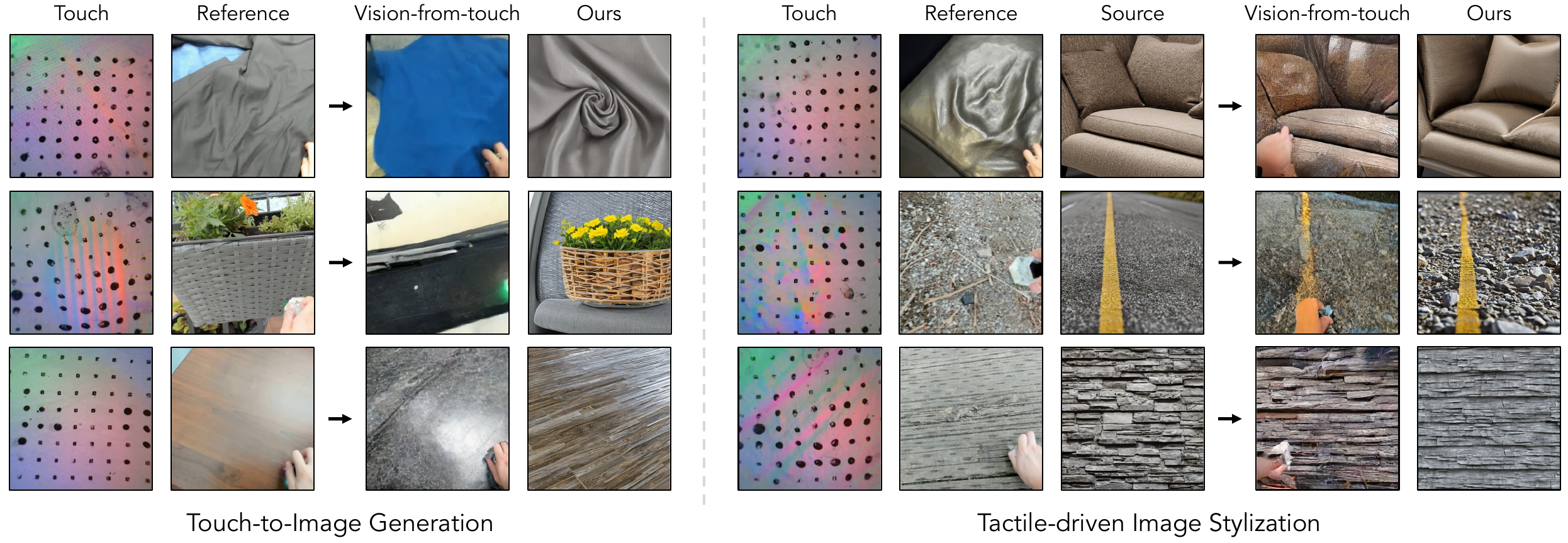}
    \caption{{\bf Zero-shot image synthesis with touch.} (Left) We generate an image of a scene given a tactile signal. (Right) We perform tactile-driven image stylization to manipulate an image to match a given touch signal. We compare our method to the state-of-the-art supervised diffusion method~\cite{yang2023generating} trained on Touch and Go. We denote ``reference" as visual images paired with the input touch in the dataset, which are not seen by the model but \textit{only} shown for the demonstration purpose. See \supparxiv{the supplement}{\cref{sec:supp_exp}} for more examples.} 
    \label{fig:synthesis}
\end{figure*}
\subsection{UniTouch representation}
First, we evaluate the quality of our learned touch features for downstream tasks: material classification and grasping stability prediction via linear probing. We freeze the learned touch embeddings and train a linear classifier on the downstream tasks for specific datasets.

\mypar{Baselines.} 
We compare our model with two recent visuo-tactile self-supervised methods for vision-based tactile sensors: VT~CMC~\cite{yang2022touch} and SSVTP~\cite{kerr2022ssvtp}. We also adopt them to our multi-dataset setup. We use the same architectures to ensure a fair comparison. 
We also compare with the supervised ImageNet~\cite{ImageNet} features, which are commonly used to represent tactile images~\cite{yuan2017shape, Calandra2018MoreTA, feel_success}.
Following~\cite{yang2022touch,feel_success, Gao_2023_CVPR}, we evaluate models' performance via accuracy metric for both downstream tasks.

\mypar{Material classification.} 
We evaluate the touch material classification task on three in-domain datasets Touch and Go, ObjectFolder 2.0, and YCB-Slide, and three out-of-domain datasets ObjectFolder 1.0, ObjectFolder Real, and SSVTP. 
It is worth noting that ObjectFolder Real and ObjectFolder 1.0 contain sensors never seen during the training.

\cref{tab:cls} shows results on linear probing. UniTouch outperforms all the baselines by a large margin, implying that our tactile representations benefit from the alignment to a well-structured embedding space trained on large-scale datasets. In addition, the consistent improvements across all datasets and sensors validate our proposed sensor-specific tokens and in-batch sampling strategy during training -- resulting in insignificant generalization gains across different sensors.

\mypar{Grasping stability prediction.} 
We follow the setting of~\cite{feel_success, Gao_2023_CVPR} to predict, from tactile input, whether a robotic gripper can successfully grasp and stably hold an object before it is lifted. Failures occur when the grasped object slips by more than 3cm. We evaluate UniTouch on three datasets: Feeling of Success, ObjectFolder 2.0, and ObjectFolder 1.0, where ObjectFolder 1.0 is an out-of-domain dataset. 

The linear probing results are shown in \cref{tab:grasp}. Our performance consistently outperforms existing baselines by a large margin. Thus, we further demonstrate that our model design and training paradigm are useful not only in computer vision but also can be generalized to robotics tasks.

\begin{table}[t!]

\centering
\resizebox{\linewidth}{!}{
\begin{tabular}{clccc}
\toprule
& \multirow{2}{*}{\textbf{Method}} & \multicolumn{3}{c}{\textbf{Retrieved Modality}}\\ 
\cmidrule{3-5}
&  &  Touch $\rightarrow$ Vision & Touch $\rightarrow$ Audio & Touch $\rightarrow$ Text \\ \midrule
 & Chance & 1.0 & 1.0 & 1.0 \\  
 \cdashlinelr{1-5} 
 \multirow{8}{*}{\shortstack[c]{Fully \\ supervised}} & CCA$^{\dagger}$ & 8.50 & 6.18 & -\\
 & PLSCA$^{\dagger}$ & 6.25 & 7.11 & -\\
 & DSCMR$^{\dagger}$ & 4.92 & 6.15 & -\\
 & DAR$^{\dagger}$ & 8.80 & 7.77 & - \\
 & CCA & 17.8 & 15.7 & 16.8\\
 & PLSCA & 16.8 & 15.9 & 18.2\\
 & DSCMR & 26.5 & 19.6 & 22.7\\
 & DAR & 32.3 & 27.8 & 31.9 \\
 \cdashlinelr{1-5} 
 \multirow{1}{*}{Zero-shot} 
 & Ours& \textbf{41.9} & \textbf{37.9} & \textbf{38.0} \\
\bottomrule
\end{tabular}}
\caption{
\textbf{Cross-modal retrieval from touch.}
We evaluate the performance using mean Average Precision (mAP) on ObjectFolder 2.0. $^{\dagger}$ denotes results from~\cite{Gao_2023_CVPR}.} %
\label{tab:retrieval_results}

\end{table}
\subsection{Zero-shot touch understanding}
We further evaluate UniTouch with zero-shot classification tasks, enabled by the emergent alignment with text during pretraining. We perform material classification and grasping prediction tasks by computing the cosine similarity between the embeddings of touch and corresponding text prompts. Class predictions are chosen based on highest scores, without training on labeled data. To the best of our knowledge, there are no other baselines that can perform zero-shot touch understanding in our manner.

\mypar{Material classification.}
We conduct zero-shot material classification by prompting the model with ``This feels like [CLS]", where [CLS] is the name of the material. We show our zero-shot performance in the last row of \cref{tab:cls}. Our zero-shot method shows a comparable performance against several supervised methods, which not only indicates a strong tactile representation that is well-aligned with the text but also shows that off-the-shelf models trained for other modalities can be used to successfully solve touch sensing tasks.

\mypar{Grasping stability prediction.}
Similarly, we perform the zero-shot grasping stability prediction task by using text prompts like ``the object is lifted in the air" and ``"the object is falling on the ground". \cref{tab:grasp} shows that we are comparable to some of the supervised methods, demonstrating the capabilities of aligning touch and text can be extended to robotics tasks, which may be out of the training scope of the vision language model like CLIP with appropriate prompting. This may come from the fact that we link the touch of the successful grasps to the robot's action of lifting objects while failed grasps as those falling. We found consistent performance in both in and out-of-distribution datasets, demonstrating the generalization capability of this link.

\begin{figure*}[t]
    \centering
    \upvspacefig
    
    \includegraphics[width=\linewidth]{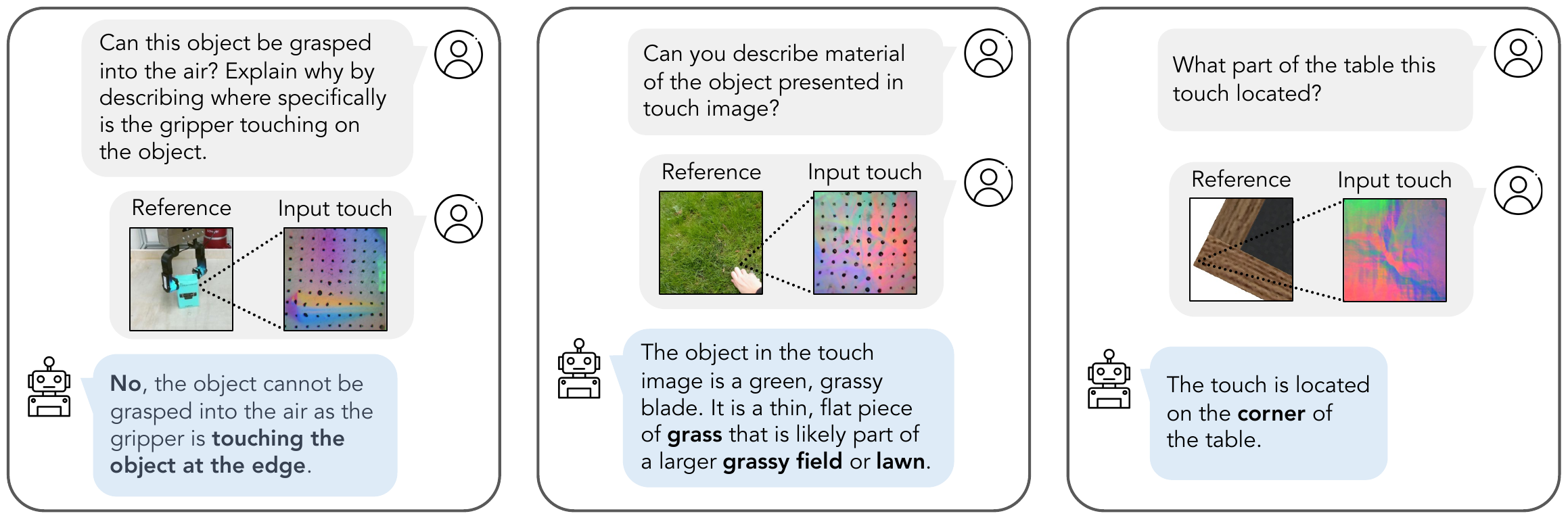}
    \caption{{\bf Touch-LLM.} Our Touch-LLM can conduct a series of tactile question-answer tasks such as robot grasping stability prediction, contact localization, and touch image captioning. We also show ``reference" visual images paired with the input touch, for better demonstration. See \supparxiv{the supplement}{\cref{sec:supp_exp}} for more examples.} 
    \label{fig:touch-llm}
\end{figure*}
\begin{table}[!t]
    \footnotesize
    \centering

    \setlength{\tabcolsep}{1mm}{
    \begin{tabular}{llccc}
    \toprule
    & \multirow{2}{*}{\textbf{Method}} & \multicolumn{3}{c}{\textbf{Evaluation Metrics}} \\ \cmidrule{3-5} 
    & & CVTP ($\uparrow$) & Material ($\uparrow$) & FID ($\downarrow$)\\
    \midrule
    \multirow{3}{*}{Supervised} & Pix2Pix~\cite{pix2pix2017} & 0.09 & 0.15 & 136.4\\
    & VisGel~\cite{Li2019ConnectingTA} & 0.10 & 0.20 & 128.3\\
    & Vision-from-touch ~\cite{yang2023generating} & 0.35 & 0.25 & \textbf{81.2}\\ 
    \cdashlinelr{1-5} 
    Zero-shot & Ours &  \textbf{0.56} & \textbf{0.31} & 103.11\\ 
    \bottomrule
\end{tabular}}
\caption{ \textbf{Zero-shot touch-to-image generation on \emph{Touch and Go}.}}
\label{tab:touch2img}
\end{table}
\subsection{Cross-modal retrieval with touch}

We conduct cross-modal retrieval to evaluate the alignment of our touch embeddings to those of other modalities. Given a touch image, we aim to identify the corresponding vision, text, and audio describing the same point of contact.

\mypar{Experimental setup.} We evaluate on ObjectFolder 2.0 cross-sensory retrieval benchmark~\cite{Gao_2023_CVPR}.
Following~\cite{Gao_2023_CVPR}, we treat points from the same object as positive samples and evaluate using mAP. To evaluate touch-to-text retrieval, we annotated text descriptions that depict the contact point of the object from its visual input, serving as paired ground-truth text. We obtain the retrieval result by ranking the cosine similarity between an input touch and other modalities. Given that our method is not trained with paired audio or text data, we consider its performance in these two modalities as a demonstration of zero-shot learning.

\mypar{Baselines.} We compare our method with several established baselines, including Canonical Correlation Analysis (CCA)~\cite{Hotelling1936RelationsBT}, Partial Least Squares (PLSCA)~\cite{Jong2001CanonicalPL}, Deep Aligned Representations (DAR)~\cite{Aytar2017SeeHA}, and Deep Supervised Cross-Modal Retrieval (DSCMR)~\cite{Zhen2019DeepSC}.

\mypar{Results.} 
UniTouch achieves state-of-the-art performance on all three modalities and outperforms those supervised methods that are trained with paired modalities by a large margin (\cref{tab:retrieval_results}). This demonstrates our strong cross-modal ability to align touch with other modalities without the need for explicit paired training data or additional supervision.
\begin{table}[!t]
\centering

\footnotesize
\begin{tabular}{llc}
\toprule
\multirow{2}{*}{\textbf{Method}}    &   \multirow{2}{*}{\textbf{LLM}}      & \textbf{Eval} \\
\cmidrule{3-3}
& & GPT-4 Rating ($\uparrow$)\\
\midrule
BLIP-2~\cite{li2023blip}        &   Vicuna~\cite{vicuna2023} &            1.01 \\
InstructBLIP~\cite{Dai2023InstructBLIPTG}    &  Vicuna~\cite{vicuna2023} &    1.93        \\
LLaVA-1.5~\cite{liu2023improved}     &  Vicuna~\cite{vicuna2023}  &         2.33   \\
 \cdashlinelr{1-3} 
Touch-LLM (ours) &  LLaMA~\cite{touvron2023llama} &  \textbf{3.30}  \\
\bottomrule
\end{tabular}
\caption{\textbf{Touch image caption evaluation.} We evaluate our Touch-LLM and three baselines on our test cases from Touch and Go~\cite{yang2022touch}. Each model's response is rated by GPT-4 on a scale from 1 to 5.} 
\label{tab:caption}
\end{table}
\subsection{Image synthesis with touch}
In this part, we demonstrate that we can combine our touch embedding with an off-the-shelf image synthesis model easily to perform the image synthesis tasks conditioning touch images in a zero-shot manner. 
We perform two tasks: touch-to-image generation~\cite{Li2019ConnectingTA,yang2023generating} and tactile-driven image stylization~\cite{yang2022touch, yang2023generating}.
Following \cite{yang2023generating, yang2022touch}, we use three evaluation metrics: Frechet Inception Distance~(FID), Contrastive Visuo-Tactile Pre-Training~(CVTP), and material classification consistency. See \supparxiv{the supplement}{\cref{sec:supp_eval}} for details.

\mypar{Touch-to-image generation.} We aim to generate images solely from touch. We use a pretrained text-to-image diffusion model~\cite{rombach2022high}, conditioning on our touch features, and guiding the denoising process.  Compared to the state-of-the-art visuo-tactile diffusion-based model~\cite{yang2023generating}, our method generates more realistic objects that have not been previously seen in the dataset (see ~\cref{fig:synthesis}~(left)). While the images generated by~\cite{yang2023generating} not only include the sensor and the arm holding it but also closely resemble the visual images in the training set. 
\cref{tab:touch2img} shows quantitative results, where we compare with Vision-from-touch~\cite{yang2023generating}, VisGel~\cite{Li2019ConnectingTA} and Pix2Pix~\cite{pix2pix2017} on Touch and Go~\cite{yang2022touch}. 
Despite a slightly lower FID score compared to \cite{yang2023generating}, our method outperforms on the CVTP and material consistency metrics. This suggests that while our generated images are out of the distribution of Touch and Go, our approach effectively bridges vision and touch.

\mypar{Tactile-driven image stylization.}
We also manipulate an image to align with a given touch signal~\cite{yang2022touch, yang2023generating} zero shot. We achieve this by mixing the input image embedding with our conditioned touch embedding and feeding it into the pretrained diffusion model. We show qualitative results in \cref{fig:synthesis} (right), where the input image is out of the distribution of Touch and Go~\cite{yang2022touch}. We observe the supervised state-of-the-art method~\cite{yang2023generating} fails to change the visual style according to the touch images even though these are seen during the training stage.
See \supparxiv{the supplement}{\cref{sec:supp_exp}} for more details.

\begin{table}[t]
\footnotesize
\centering
\resizebox{0.9\linewidth}{!}{
\begin{tabular}{@{}lcc@{}}
\toprule
\multirow{2}{*}{\textbf{Prompt}}  & \multicolumn{2}{c}{\textbf{Datasets}} \\
\cmidrule{2-3} 
& \emph{Touch and Go} & \emph{OF 2} \\ 
\midrule
{\tt This is an image of [CLS]}& 40.7  & 34.3\\
{\tt This is a \textbf{touch} image of [CLS] } &\textbf{43.8}  & \textbf{36.8} \\
\cdashlinelr{1-3}
{\tt This looks like [CLS] }& 49.3& 41.7\\
{\tt This \textbf{feels} like [CLS]} & \textbf{52.7}  &  \textbf{43.5}\\
\cdashlinelr{1-3}
{\tt Image of [CLS]}& 48.8 & 40.3\\
{\tt \textbf{Touch} of [CLS]}& \textbf{51.2}  &  \textbf{40.9} \\ \bottomrule
\end{tabular}}
\caption{ \textbf{Prompt analysis for touch.} We evaluate our prompt designs for zero-shot material classification on Touch and Go and ObjectFolder 2.0 datasets.}
\label{tab:prompt}
\end{table}
\subsection{Touch-LLM}
Interpreting vision-based touch images, crucial for delicate tasks in fields like robotics, is challenging due to human perceptual limitations. To address this, we integrate UniTouch embedding into a large language model (LLM), leveraging its robust understanding and reasoning capabilities for touch image interpretation, and name it as Touch-LLM.
Touch-LLM is capable of a series of tactile tasks such as grasping stability prediction, touch image interpretation, tactile contact localization and \etc, most of which are non-trivial to humans, demonstrating the usefulness of combining touch with LLMs. We show some example tasks in \cref{fig:touch-llm}.

Quantitatively, we compare our model with three open-source vision-language models (VLMs): BLIP-2~\cite{li2023blip}, InstructBLIP~\cite{Dai2023InstructBLIPTG}, and LLaVA-1.5~\cite{liu2023improved} in the touch image captioning task by feeding them the same touch images and text prompts. 
We manually create captions for 400 randomly sampled RGB images from Touch and Go~\cite{yang2022touch} as the ground truth. 
Following~\cite{bitton2023visit}, we use GPT--4 to perform automatic evaluation by instructing GPT-4 to rate each model's generations on a scale of 1 to 5 given the reference response.  As shown in \cref{tab:caption}, our Touch-LLM outperforms other VLMs by a large margin, indicating that our Touch-LLM has much better understanding capabilities for touch images even with a less powerful LLM than Vicuna~\cite{vicuna2023} which used by other models. See \supparxiv{the supplement}{\cref{sec:supp_eval}} for more details.

\subsection{X-to-touch generation}
We conduct X-to-touch generation to synthesize realistic tactile images corresponding to the input modality of vision, language, and audio. \cref{fig:teaser} shows plausible and consistent tactile images generated from both the visual input and its text captioning. 
Quantitatively, we evaluate our model on Touch and Go~\cite{yang2022touch}, where we measure material classification consistency between touch images generated from vision and its corresponding language captions. Our model achieves 55.3\% consistency, illustrating the reliability of the generated results.
See \supparxiv{the supplement}{\cref{sec:supp_exp,sec:supp_eval}} for more examples and details.

\subsection{Ablation study}
\vspace{3mm}
\mypar{Learning from multiple sensors.}
\cref{tab:ablation} ablates the importance of each module design on the zero-shot material classification task with the Touch and Go dataset.
The baseline, a vanilla transformer model aligning touch embedding to a fixed vision encoder, drops performance significantly when applied to multiple sensors and datasets, \ie, from 43.1\% to 21.4\%, indicating the difficulty of the sensor domain gap. We improve the performance by 17\% by adding the sensor-specific tokens to it. Similarly, we found a 19\% by adding our sampling strategy. With our proposed batch sampling strategy and sensor-specific tokens, our model can achieve strong performance, surpassing the model trained on a single dataset, which emphasizes the significance of our proposed methods for learning a better touch representation from multiple sensors. 
We argue that this is because sensor-specific embeddings help distinguish hard samples from different sensors while sampling strategy helps identify hard negatives within the same sensor in the training. Combining these, we can tackle inter-sensor and intra-sensor hard samples thus obtaining the performance boost.

\mypar{Language prompting for touch.}
We explore how language prompting can help with understanding touch, the first endeavor in this domain.
Given that vision captures more global and semantic information, and touch focuses on material properties, texture, and microgeometry, directly adopting prompts from vision-language works may not yield satisfactory results. We design touch-specific prompt templates by adopting the common prompts from vision-language works and replacing with words related to haptics, i.e., changing ``{\tt image}" to ``{\tt touch image}" and ``{\tt look like}" to ``{\tt feel like}" (see \cref{tab:prompt}). 
We evaluate them using the zero-shot material classification task on Touch and Go and ObjectFolder 2.0. We empirically found that our prompts can significantly improve the performance, indicating that language can indeed understand touch. We suspect this phenomenon may be due to the design of visuo-tactile datasets, which feature human or robotic touch actions, thus enabling the model to associate tactile images with these actions.

\begin{table}[!t]
\footnotesize
\centering
\resizebox{\linewidth}{!}{
\begin{tabular}{llc}
\toprule
\multirow{2}{*}{\textbf{Method}} & \multirow{2}{*}{\shortstack[c]{\textbf{Pretrain} \\ \textbf{Data}}} & \textbf{Eval} \\ \cmidrule{3-3}
&  & \emph{Touch and Go}\\ \midrule
Chance & -- & 16.7 \\ 
Baseline & \emph{Touch and Go} & 43.1 \\
Baseline & All & 21.4 \\
Baseline + sensor token & All & 38.1 \\
Baseline + sample& All & 40.3 \\
Baseline + sensor token + sample & All & \textbf{52.7} \\
\bottomrule
\end{tabular}}
\caption{\textbf{Ablation study.} We ablate the effectiveness of each of our proposed contributions via the zero-shot material classification.}
\label{tab:ablation}
\end{table}

\section{Discussion}
We introduced \textit{UniTouch}, a unified multimodal tactile representation for vision-based tactile sensors. 
To achieve this, we align our touch embedding to a shared multimodal embedding space using contrastive learning. 
We further introduce sensor-specific tokens that enables learning from different sensors all at once. UniTouch unifies many existing tactile sensing tasks and significantly expands the range of tasks for touch sensing.
Nonetheless, the field of multimodal (foundational) model is admittedly still young. Agents, like ourselves, leverage complementary strengths of multi-sensory observations, incorporating all five senses in everyday tasks. With that goal in mind, we see our work as a concrete step towards that direction, opening new avenues for multimodal touch experience beyond vision and touch and integrating tactile sensing into multimodal foundation models. 

\mypar{Limitations.} As the full range of tactile sensors exhibit differing output formats (e.g. image, barometric signals, force), we limit our scope to vision-based tactile sensors. Scaling up our training strategy is key to further integrate emerging tactile sensors in the future. In addition, like other multimodal foundational models, our representation is ``black-box'', which does not easily for interpretability in the space, where one may benefit from explainability.

\mypar{Acknowledgements.} 
We thank Jiacheng Zhang, Shaokai Wu and Chenyang Ma for the helpful discussions and feedbacks on our manuscript. This work is supported by NSF 2112562 Athena AI Institute and Sony Research.
{
    \small
    \bibliographystyle{ieeenat_fullname}
    \bibliography{main}
}

\clearpage

\supparxiv{
\setcounter{page}{1}
\maketitlesupplementary
}{}

\appendix
\renewcommand{\thesection}{A.\arabic{section}}
\setcounter{section}{0}

\section{Datasets and Metrics}
\label{sec:supp_dataset}
We provide more details of datasets used in our paper, all of which are publicly available.
\vspace{1mm}
\mypar{Touch and Go~\cite{yang2022touch}.} The Touch and Go dataset is a recent, real-world visuo-tactile dataset featuring human interactions with various objects in both indoor and outdoor environments using a GelSight tactile sensor. It comprises 13,900 instances of touch across approximately 4,000 distinct object instances and 20 types of materials. Since it is the only real-world in-the-wild dataset, we apply it to multiple tasks including material classification, image synthesis with touch, Touch LLM, and X-to-touch generation. We use the official train/test split of~\cite{yang2022touch} where the dataset is split by touches, not by frames to avoid similar touch images between the train and test set. For Touch-LLM and X-to-touch applications, we label 400 visual images by asking turkers to provide their captioning to describe the object, touch feeling, and texture from it.

\mypar{The feeling of success~\cite{feel_success}.} The Feeling of Success is a robot-collected visuo-tactile dataset of robots grasping objects on a tabletop. The tactile images are all captured by GelSight tactile sensors. It contains 9.3k paired vision and touch images. We apply this dataset to robotic grasping stability predictions. As there is no official split of train/val/test, following~\cite{yang2022touch, Gao_2023_CVPR}, we split the dataset by objects in the ratio of 8:1:1.

\mypar{YCB-Slide~\cite{suresh2022midastouch}.}
The YCB-Slide dataset comprises DIGIT sliding interactions on YCB objects. The dataset is in the video format where we take all 180k frames for our experiments. The dataset contains 10 YCB objects including a sugar box, a tomato soup can, a mustard bottle, a bleach cleanser, a mug, a power drill, scissors, an adjustable wrench, a hammer, and a baseball. While the tactile images are collected via sliding interaction, the visual input is generated by simulation of the YCB objects. In our experiment, we treat each of the objects as an individual material and our goal is to classify 10 classes. We apply this dataset to material classification.

\mypar{ObjectFolder 1.0~\cite{gao2021ObjectFolder}.} 
The ObjectFolder 1.0 dataset is a simulation dataset containing 3D models of 100 objects from online repositories. The touch images are simulated by TACTO simulators. As the raw dataset is a 3D model with infinite points, we randomly sample 200 points for each object. We apply this dataset to material classification and grasping stability prediction experiments. It is worth noting that for grasping stability prediction experiments, we select 6 objects suitable for grasping following their setting and achieve relatively balanced successful and failure outcomes for grasping. Following~\cite{gao2021ObjectFolder}, all materials can be categorized into 7 material categories including wood, steel, polycarbonate, plastic, iron, ceramic, and glass. These categories are also applied to ObjectFolder 2.0 and ObjectFolder Real datasets.

\mypar{ObjectFolder 2.0~\cite{gao2022ObjectFolderV2}.} The ObjectFolder 2.0 dataset extends~\cite{gao2021ObjectFolder} to 1000 objects and improves the acoustic and tactile simulation pipelines to render more realistic multisensory data. For the tactile simulation, it utilizes the Taxim simulator instead of TACTO. Similar to the preprocessing of ObjectFolder 1.0, we sample 200 points for each object. To avoid overlapping with~\cite{gao2021ObjectFolder}, we only take the 101-1000 objects. We apply this dataset to material classification, cross-modal retrieval, robot grasping stability prediction, and Touch-LLM. For cross-modal retrieval and Touch-LLM tasks, we annotate text descriptions that depict the contact point of the object from its visual input, \eg ``{\tt The corner of a wooden table.}" 

\mypar{ObjectFolder Real~\cite{Gao_2023_CVPR}.} ObjectFolder Real is an object-centric multimodal dataset containing 100 real-world household objects. The touch images are captured by the GelSlim tactile sensor. Similarly, we sample 200 points for each object thus containing in total of 20k visuo-tactile pairs. We apply this dataset to a material classification task, which is considered an out-of-domain dataset.

\mypar{SSVTP~\cite{kerr2022ssvtp}.} SSVTP dataset is a recent human-collected visuo-tactile dataset containing 4.9k paired visuo-tactile images. The touch images are collected via the DIGIT tactile sensor. The objects in this dataset are mainly from garments but also contain materials of metal. We apply this dataset to material classification. As the dataset does not contain material labels, we annotate material labels from the visual images. In total, we classify all images into 6 material categories including cotton, metal, denim fabric, plastic, wood, and nylon.

\section{Implementation Details}
We show more implementation details in this section.
\mypar{Image synthesis with touch.} We used a pretrained stable diffusion-2.1 unclip~\cite{rombach2022high} to perform zero-shot touch-to-image generation by replacing the text condition with our aligned UniTouch embedding. Specifically, we keep the simple text {\tt "high quality"} as the condition while using our touch embedding as an additional condition. We use DDIM sampler~\cite{song2020denoising} with a guidance scale of 9 and denoising steps of 50. Additionally, we set an embedding strength of 0.75 for our touch embedding condition. Synthesized images are at the resolution of 768$\times$768. 

As for tactile-driven image stylization, similarly, we still keep the simple text {\tt "high quality"} as the condition. However, we use both touch and image embeddings as extra conditions to conduct image stylization. We perform a linear combination of touch and image embeddings, the weights for touch and image are set to 0.3 and 0.7 respectively. We use DDIM sampler~\cite{song2020denoising} with a guidance scale of 9 and denoising steps of 50. The strength for linear combination embedding is set to 1 and edited images are at the resolution of 768$\times$768. 

\mypar{Touch-LLM.}
We adapt our model from~\cite{gao2023llama, zhang2023llama}, which leverages an adapter to connect our touch encoder and an open-source large language model LLaMA~\cite{touvron2023llama}. We replace RGB image embedding with our aligned UniTouch embedding. Concretely, we denote the global touch feature encoded by our touch encoder as $F_{T} \in \mathbb{R}^{1 \times C_{T}}$, where $C_{T}$ is the dimension of the touch embedding. Inspired by prior work~\cite{gao2023llama, zhang2023llama}, we use a projector $f$, which encodes $F_{T}$ to have the same dimension as the token embedding in LLaMA~\cite{touvron2023llama}:
\begin{equation}
    F^{\prime}_{T} = f\left(F_{T}\right)\text{.}
\end{equation}
Then we repeat $F^{\prime}_{T}$ and add it to all text tokens across all layers in language model LLaMA~\cite{touvron2023llama} with a zero-initialized learnable gate function:
\begin{equation}
    T_{j}^{q} = h_{\text{zero}} \cdot F^{\prime}_{T} + T_{j}^{q},
\end{equation}
where $j$ and $q$ denotes the layer and sequence index respectively, $T_{j}^{q}$ is the text token embedding, and $h_{\text{zero}}$ is the zero-initialized learnable gate function. In our experiments, we use pretrained $h_{\text{zero}}$, and plug our UniTouch embedding in. 
\mypar{X-to-touch generation} We conduct our X-to-touch generation model based on stable diffusion. While most existing multimodal tactile datasets only contain vision and touch, we first train an image-to-touch diffusion model and we are able to conduct text-to-touch and audio-to-touch \textit{zero shot} by replacing the image conditioning as they are already aligned. We use the Adam optimizer with a base learning rate of 1e-6. 
Models are all trained with 30 iterations using the above learning rate policy. We train our model with a batch size of 48 on 4 RTX A40 GPUs. Since we want to use the aligned condition embeddings, the conditional model is frozen during training. The condition embeddings are integrated into the model using cross-attention. We use the frozen, pretrained VQGAN to obtain our latent representation, with a spatial dimension of 64×64. During the inference, we conducted the denoising process for 200 steps and set the guidance scale s = 7.5.

\section{Evaluation Details}
\label{sec:supp_eval}
\vspace{5mm}
\mypar{Touch-to-image generation}
Following~\cite{yang2023generating}, we use three evaluation metrics of Frechet Inception Distance (FID), Contrastive Visuo-Tactile Pre-Training (CVTP), and Material Classification Consistency. FID is a standard evaluation metric in image synthesis that compares the distribution of real and generated image activations using a trained network. CVTP~\cite{yang2023generating} is a metric similar to CLIP but measures the cosine similarity between the visual and tactile embeddings learned for the generated images and conditioned tactile signals, which used an off-the-shelf network. Material classification consistency~\cite{yang2023generating} uses a material classifier to categorize the predicted and ground truth images and measure the rate at which they agree, where we use CLIP as the zero-shot material classifier by feeding the prompt of {\tt "material of [CLS]"}. 

\mypar{Touch-LLM.} 

\begin{figure*}[!t]
    \centering
    \upvspacefig
    \includegraphics[width=\textwidth]{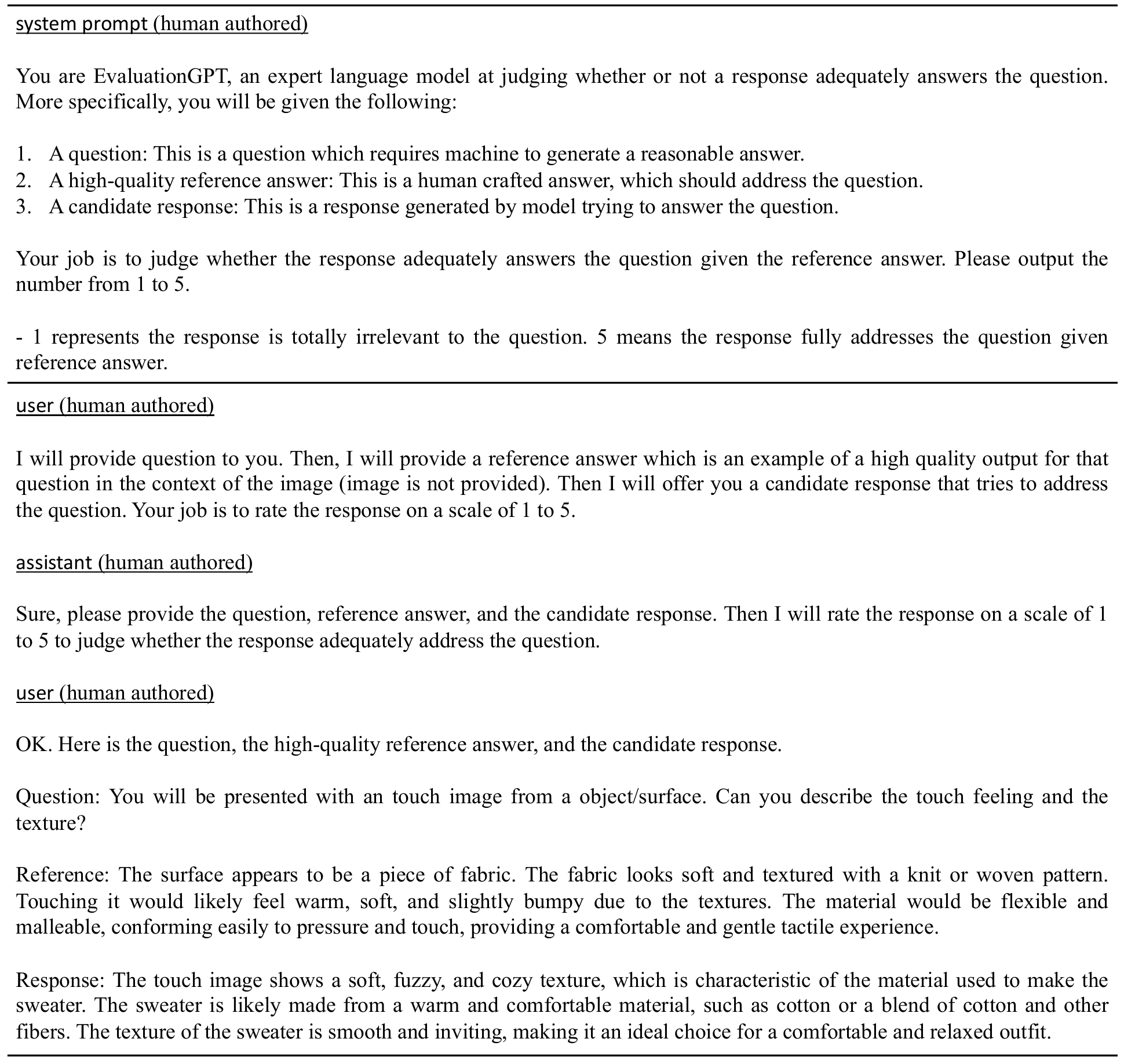}
    
    \caption{{\bf GPT-4 evaluation template.} We use this template to instruct GPT-4 for automatic evaluation of our Touch-LLM and other selected open-source VLM baselines.  } 
    \label{fig:gpt-4 eval}
\end{figure*}

We feed each vision language model (including our Touch-LLM) with a touch image and text prompt: {\tt "You will be presented with a touch image from an object/surface. Can you describe the touch feeling and the texture?"}. In the end, we use GPT-4 to perform the automatic evaluation for each model following prior work~\cite{bitton2023visit}. Specifically, we provide GPT-4 with: 1) a system prompt describing the desired evaluation behavior; 2) the question; and 3) a human-crafted reference response; 4) each model's generation result (more details see supp.). We instruct GPT-4 to rate each model's generations on a scale of 1 to 5 given the reference response. The template is shown in \cref{fig:gpt-4 eval}.

\mypar{X-to-touch.} We test the effectiveness of the x-to-touch model on the Touch and Go dataset, which is the only real-world dataset that contains objects and scenes in the wild. As the objects in this dataset are closely related to the material properties, we measure the material classification consistency between different touches generated from different modalities. We use our UniTouch embedding as the off-the-shelf zero-shot material classifier. For quantitative results for text-to-touch generation, we use the 400 human-labeled text captions as the input. For audio-to-touch generation, as there is no impact sound correlated to this dataset, we manually select audios from ObjectFolder 2.0 as the input that have the same material properties or geometry with the visual image for qualitative evaluations, as shown in \cref{fig:x2touch_supp}. 

\section{Additional Experiments}
\label{sec:supp_exp}
\vspace{5mm}

\begin{figure}[t]
    \centering
    \upvspacefig
    \includegraphics[width=0.8\linewidth]{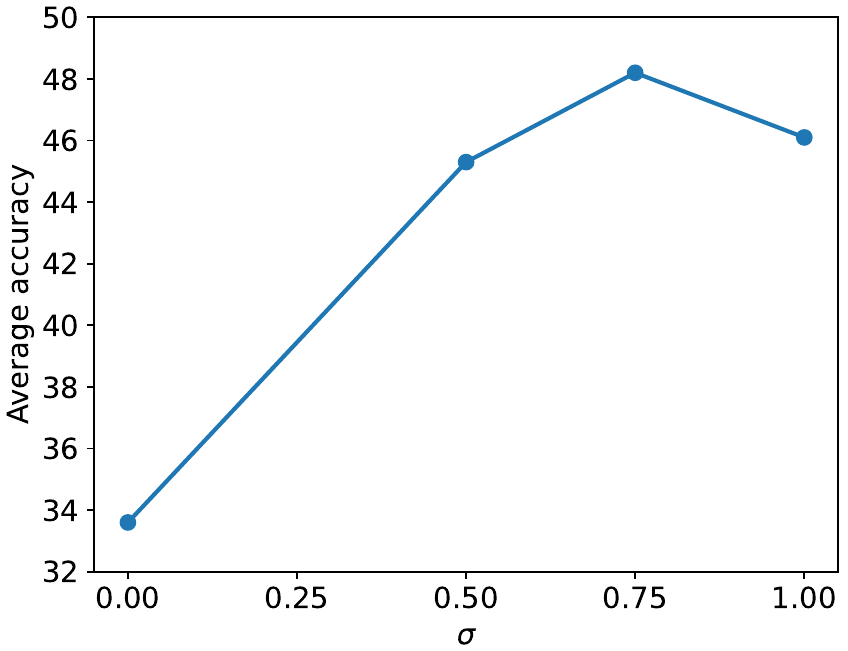}
    
    \caption{{\bf Effect of $\sigma$ for in-batch sampling.} We compare the average zero-shot material classification accuracy from six datasets using different $\sigma$ of 0, 0.5, 0.75, 1.  } 
    \label{fig:mix_rate}
\end{figure}

\mypar{In-batch sampling mix rate selection.} We evaluate different choices of $\sigma$ for in-batch sampling, where $\sigma$ denotes the percentage of the data that comes from the same dataset while the rest from others. We set $\sigma$ to $\{0, 0.5, 0.75, 1.0\}$ and evaluate their zero-shot material classification performance on all six datasets, as shown in \cref{fig:mix_rate}. We observe that if we select $\sigma = 0$, the ability to distinguish between intra-sensor samples is significantly undermined thus leading to inferior performance. As the $\sigma$ is increasing, the model is able to better distinguish between intra-sensor samples. In the extreme case when $\sigma = 1.0$ where all samples come from the same dataset, the model will have no exposure to the inter-class negatives. We observe that the performance in this case is actually decreasing. This demonstrates the effectiveness of design to balance between inter-sensor and intra-sensor negatives. We empirically found that selecting $\sigma = 0.75$ obtains a good trade-off between these factors.

\begin{figure*}[t]
    \upvspacefig

    \centering
    \includegraphics[width=0.85\linewidth]{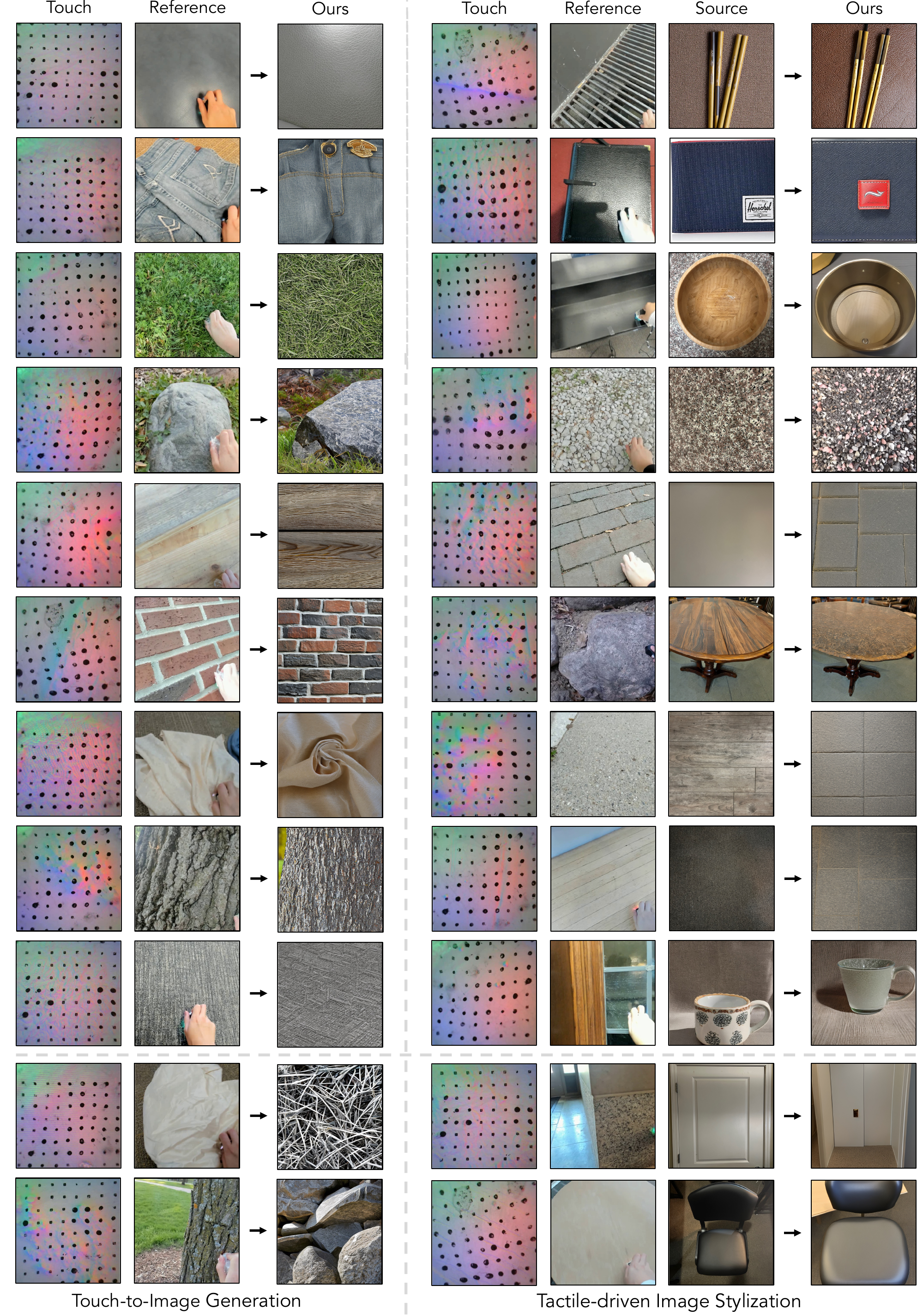}
    \caption{{\bf More examples of zero-shot image synthesis with touch.} (Left) We generate an image of a scene given a tactile signal. (Right) We perform tactile-driven image stylization to manipulate an image to match a given touch signal. We denote ``reference" as visual images paired with the input touch in the dataset, which are not seen by the model but \textit{only} shown for demonstration purposes. The last two rows are failure cases.} 
    \label{fig:synthesis_supp}
\end{figure*}

\mypar{Image synthesis with touch.} We leverage our aligned UniTouch embedding and pretrained text-to-image stable diffusion model~\cite{rombach2022high} to generate more qualitative results of touch-to-image generation and tactile-driven image stylization as presented in \cref{fig:synthesis_supp}. It shows that our UniTouch embedding can guide image synthesis successfully in a zero-shot manner. 

\begin{figure*}[t]
    \centering
    \upvspacefig
    
    \includegraphics[width=\linewidth]{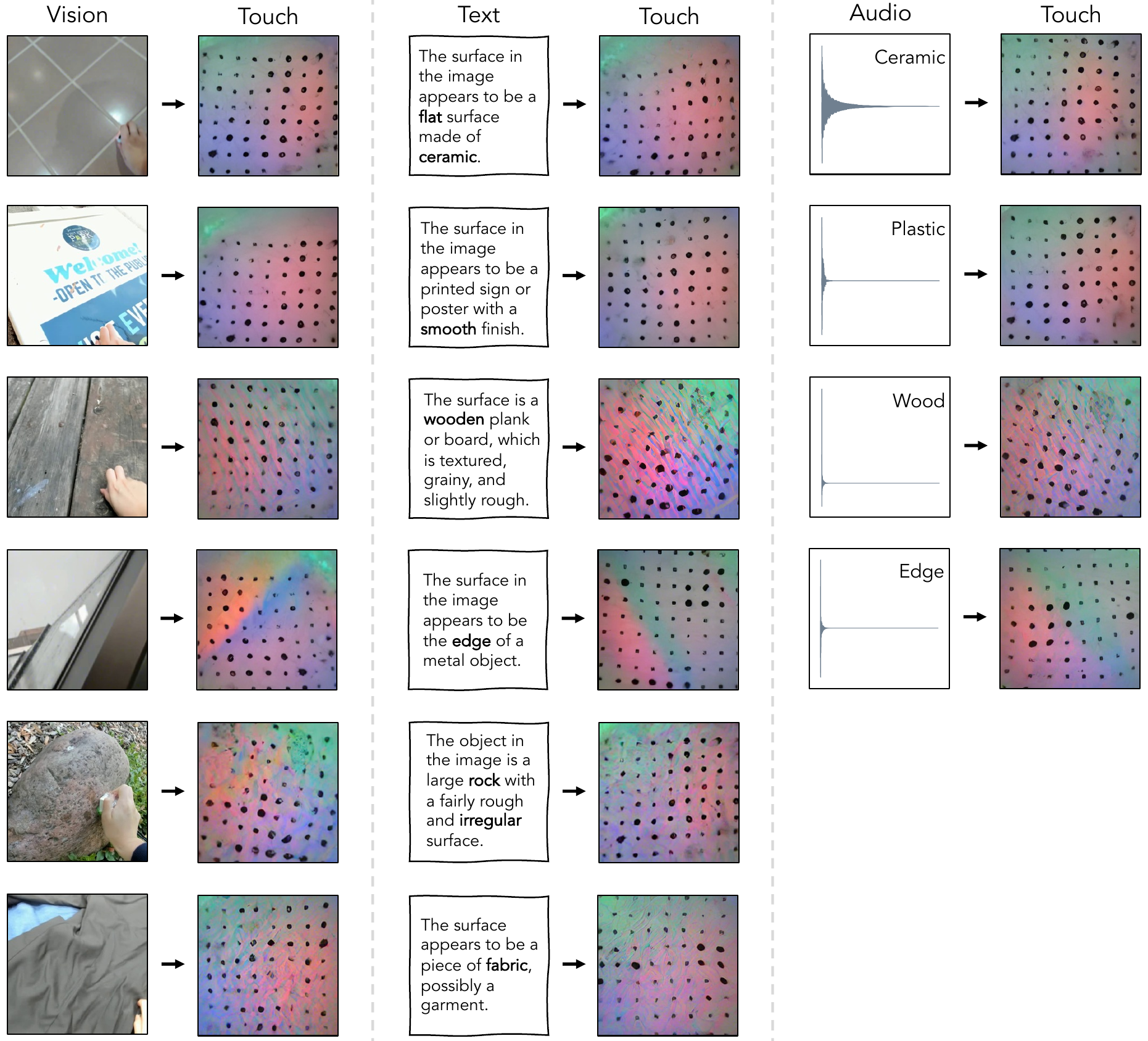}
    \caption{{\bf More examples for X-to-touch generation.} We show more examples of x-to-touch generations on the Touch and Go~\cite{yang2022touch} dataset. We manually select audios from ObjectFolder 2.0~\cite{gao2022ObjectFolderV2} matching the vision input. Since the overlapping material categories between~\cite{gao2022ObjectFolderV2} and~\cite{yang2022touch} are limited and~\cite{gao2022ObjectFolderV2} only contains rigid objects, impact sound for materials like stone and cloth can not be found. }
    \label{fig:x2touch_supp}
\end{figure*}

\mypar{X-to-touch generation.}
We show more examples of X-to-touch generations on the Touch and Go~\cite{yang2022touch} dataset in \cref{fig:x2touch_supp}, where we generate touch images using image, text, and audio.

\begin{figure*}[t]
    \centering
    \upvspacefig
    
    \includegraphics[width=0.9\linewidth]{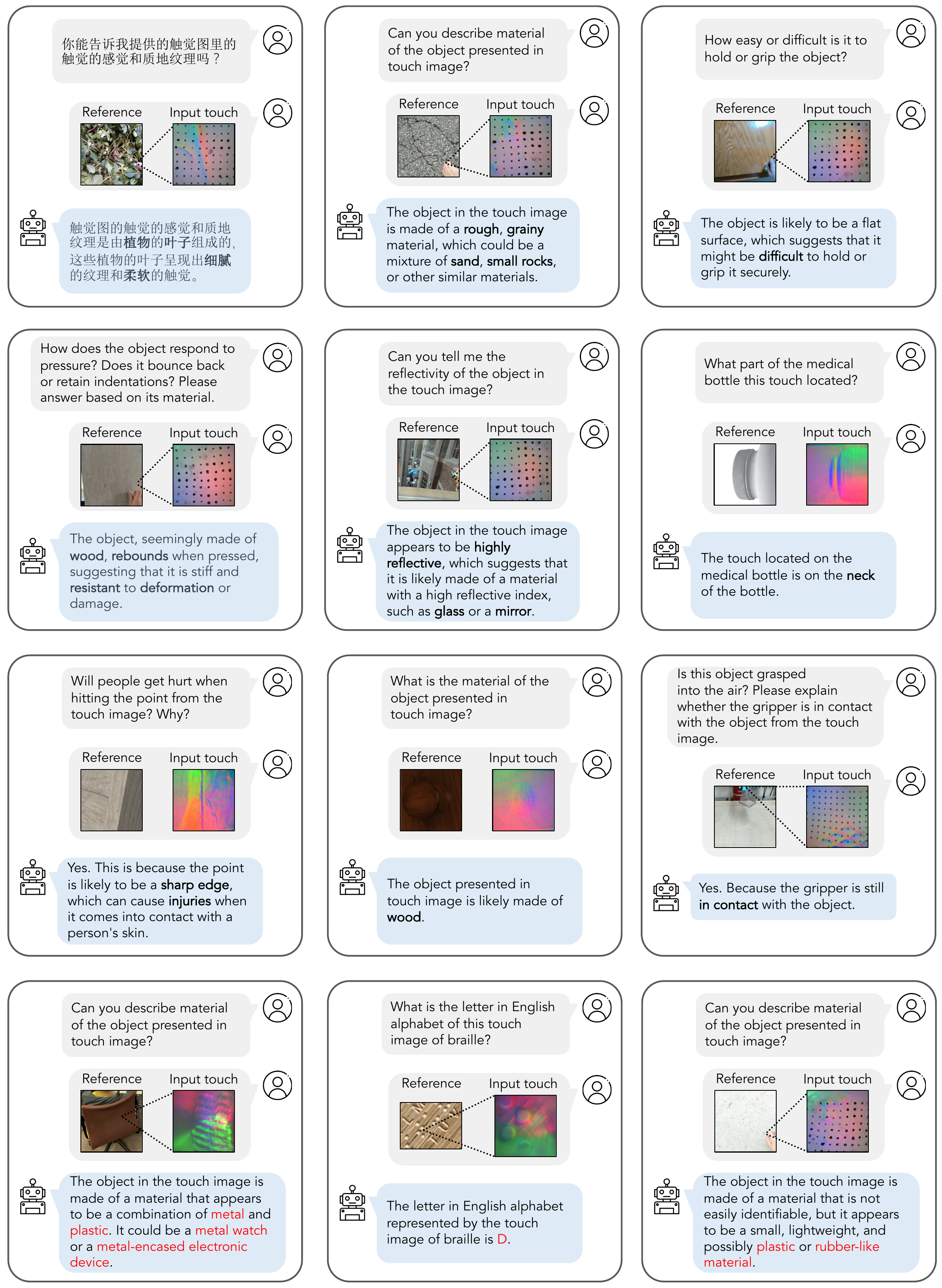}
    \caption{{\bf More examples of Touch-LLM.} We show more question-and-answering examples for touch images using our Touch-LLM. We denote ``reference" as visual images paired with the input touch in the dataset, which are not seen by the model but \textit{only} shown for demonstration purposes. The last row is the failure case. Incorrect portion is highlighted in red.} 
    \label{fig:llm-supp}
\end{figure*}

\mypar{Touch-LLM.} We show more touch image question answering examples in \cref{fig:llm-supp}.


\end{document}